# FUZZY RULE INTERPOLATION METHODS AND FRI TOOLBOX

[1]MAEN ALZUBI, [2]ZSOLT CSABA JOHANYÁK, [3]SZILVESZTER KOVÁCS,

[1, 3] Department of Information Technology, University of Miskolc, H-3515 Miskolc, Hungary

[2]Department of Information Technology, John Von Neumann University, Hungary

E-mail: [1]alzubi@iit.uni-miskolc.hu, [2]johanyak.csaba@gamf.uni-neumann.hu, [3]szkovacs@iit.uni-miskolc.hu

**ABSTRACT**

FRI methods are less popular in the practical application domain. One possible reason is the missing common framework. There are many FRI methods developed independently, having different interpolation concepts and features. One trial for setting up a common FRI framework was the MATLAB FRI Toolbox, developed by Johanyák et. al. in 2006. The goals of this paper are divided as follows: firstly, to present a brief introduction of the FRI methods. Secondly, to introduce a brief description of the refreshed and extended version of the original FRI Toolbox. And thirdly, to use different unified numerical benchmark examples to evaluate and analyze the Fuzzy Rule Interpolation Techniques (FRI) (KH, KH Stabilized, MACI, IMUL, CRF, VKK, GM, FRIPOC, LESFRI, and SCALEMOVE), that will be classified and compared based on different features by following the abnormality and linearity conditions [15].

## 1. INTRODUCTION

Former popularity of fuzzy control application was derived from the simple human-readable knowledge representation of fuzzy rules and the simple heuristic way of the control surface definition. Using fuzzy sets as linguistic terms and defining a control surface by fuzzy rules as overlapping fuzzy points was a simple way to express and implement a heuristic control strategy.

On the other hand, the heuristic definition of the fuzzy rule base in a higher dimensional problem is a challenging task. The traditional fuzzy systems, [1], [2] were implemented based on defining a complete rule base. In the complete fuzzy rule base, we have to consider all the possible rule base. The fuzzy reasoning is based on rule firing strengths i.e. rule matching calculated from the t-norm of fuzzy sets, the required rule base size is exponential with the number of the input dimensions.

However, in case the complete fuzzy rule base cannot be obtained for any reason (e.g. lack of expert knowledge base or no overlapping of fuzzy sets), then the classical reasoning methods cannot offer the desired conclusion. That happens because there may be a new observation that is not covered directly by any of the current fuzzy rule base. In this case, the fuzzy rules are considered as a sparse rule-base. There are several application areas such as control system, intrusion detection system and etc., requested a conclusion for each observation. This case the classical reasoning methods could face the problem of missing conclusion for some of the observations.

Alternative fuzzy reasoning solutions, i.e. Fuzzy Rule Interpolation (FRI) methods can release the need for the complete rule-base by replacing the rule matching reasoning concept with fuzzy interpolating function. The Fuzzy Rule Interpolation (FRI) methods were produced to handle the case of sparse rule-base. FRI methods are suitable to produce a conclusion even if some observations are not covered directly by the fuzzy rules. Therefore, using the FRI methods there is no need to have a complete fuzzy rules. The most significant fuzzy rules are enough to generate the desired conclusion.

The goals of this paper are divided as follows: firstly, to present a brief introduction of the FRI methods. Secondly, to introduce a brief description of the refreshed and extended version of the original FRI Toolbox. And thirdly, to use different unified numerical benchmark examples to evaluate and analyze of the Fuzzy Rule Interpolation Techniques (FRI) (KH, KH Stabilized, MACI, IMUL, CRF, VKK, GM, FRIPOC, LESFRI, and SCALEMOVE) that will be classified and compared based on the abnormality and linearity conditions.

The rest of the paper is organized as follows: Section (2) provides a brief review of the basic definitions of the classical reasoning and interpolative reasoning methods. Section (3) introduces an overview about enumeration of some of the implemented FRI methods. Description of the renewed and extended version of the original FRI toolbox is presented in section (4) and a set of some numerical examples of implemented FRI methods are presented in section (5). FRI methods results are discussed in section (6). Finally, section (7) concludes the paper.

## 2. PRELIMINARIES

This section provides a brief overview of the basic definitions of the complete fuzzy rule base and sparse rules. It also briefly introduces the description of the interpolative reasoning concept.

### 2.1 Complete and Incomplete Rule Bases

Let us take into consideration two numerical variables X and Y described on the universe R of real numbers, and F is a set in the fuzzy sets of R. We assume the fuzzy sets $A_i$ in F are defined, $1 \leq i \leq n$, such that: $A_1 \preceq A_2 ... \preceq A_i \preceq A_{i+1} ... \preceq A_n$, for a given order $\preceq$ on F. We also suppose that we are given fuzzy sets $B_i$ in F, $1 \leq i \leq n$, which are also ordered according to $\preceq$.

According to the definitions in [12], [13], the fuzzy functions are described by the fuzzy relations between the fuzzy sets of the inputs $A_i$ and outputs $B_i$. The fuzzy rule base could be characterized and represented based on this relation. The classical reasoning methods, such as Mamdani and Sugeno [1], [2] follow that relation which require to define all the fuzzy rule base relations between the inputs and outputs, in addition, to define the overlapping between them to get the desired conclusion. Figure (1) describes the complete fuzzy rule base between two dimensions antecedents and single consequent, the observations (x1) and (x2) are matching with the fuzzy rules 1,2,4 and 5, thus, the conclusion could be computed based on one of the classical fuzzy reasoning methods, like the Zadeh-Mamdani max-min Compositional Rule of Inference (CRI).

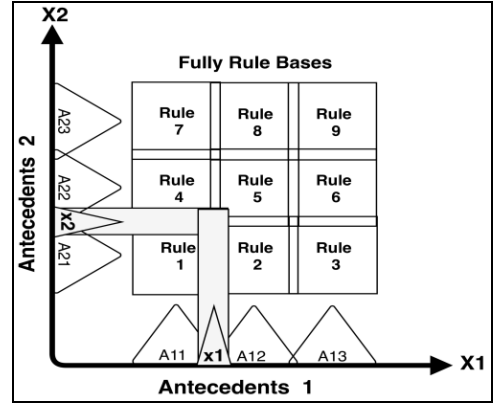

*Figure 1: Complete Fuzzy Rule Base*

Regarding the sparse rule-base (incomplete rule-bases) systems where fuzzy rules are of the type: (Ri): "if X is Ai then Y is Bi". The sparsity means there is no overlapping between the observation and any of the fuzzy rules (do not cover the input space F), where there exist inputs A* such that $\exists_i / A_i \preceq A^* \preceq A_{i+1}$. The aim of a fuzzy interpolation method is to provide the conclusion corresponding to the observation A* by considering only the two rules $R_i$ and $R_{i+1}$ when $A_i \preceq A^* \preceq A_{i+1}$.

Figure (2) describes the issue, where the observations $x_{1.1}$ and $x_{1.2}$ refer to the first input (antecedent 1), the observations $x_{2.1}$ and $x_{2.2}$ refer to the second input (antecedent 2). These observations are described two different types of issues in classical reasoning. The observations $x_{1.1}$ and $x_{2.1}$ are not overlapped with any of the rules of the rule-base, while, the observations $x_{1.2}$ and $x_{2.2}$ hit spaces in the universe of discourse, there are no linguistic terms defined, hence no overlapping rule can exist.

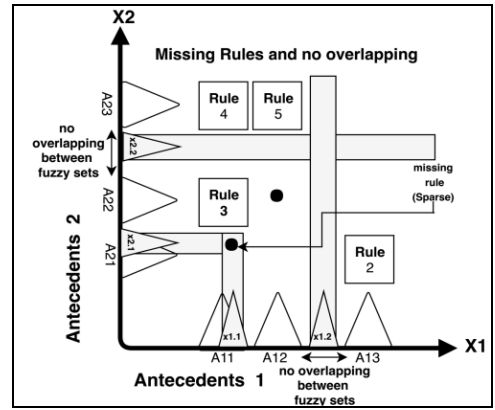

*Figure 2: The Incomplete Rule Base (Sparse Fuzzy Rule Base And No Overlapping Fuzzy Sets)*

## 2.2 Notation of FRI

According to the definition of the fuzzy function, the fuzzy space can be described by the mapping between antecedents and consequents fuzzy sets $L^X$ and $L^Y$ via $f: L^X \rightarrow L^Y$. This leads to the main idea of the fuzzy rule interpolation methods which is finding a suitable fuzzy interpolating function.

These functions could be able to produce a conclusion directly even if the rule base is sparse, and there is no overlapping between the observation and any of the fuzzy rules.

Many of the fuzzy rule interpolation (FRI) methods following the notion in [3], [14], [15] which describe the relation between two fuzzy rule base, these fuzzy sets must be adjacent convex and normal (CNF) and partially ordered fuzzy sets. Where the ordering is defined as $A_1$, is said to be "less than" $A_2$, for all $A_1$, $A_2$ sets in a given fuzzy partition. the ordering of the fuzzy set $A_1$ and $A_2$, denoted by $A1 \prec A2$, if $\forall \alpha \in [0,1]$, the following condition hold:

$$\inf(A_{1\alpha}) < \inf(A_{2\alpha}), \sup(A_{1\alpha}) < \sup(A_{2\alpha}),$$

Where the "inf" denotes the infimum and "sup" refers the supremum of the $(A_{1\alpha})$, $(A_{2\alpha})$ fuzzy sets. For simplicity, suppose that two fuzzy rules are given:

*If X is $A_1$ then Y is $B_1$*
*If X is $A_2$ then Y is $B_2$*

Where the fuzzy rules are described by $A_1 \Rightarrow B_1$ and $A_2 \Rightarrow B_2$. Also, that rules in a given rule base are arranged with respect to a partial ordering among the convex and normal fuzzy sets (CNF sets) of the antecedents, consequent and observation. For the above two rules, this means that:

$$A_1 \prec A^* \prec A_2 \ \wedge B_1 \prec B_2$$

Figure (3) illustrates the simplest form to describe two flanking rules of the fuzzy sets, where the shape of the fuzzy sets membership functions remained restricted to trapezoidal, the figure shows the main points (variables) of the fuzzy sets to be applied for determining the conclusion in most FRI methods.

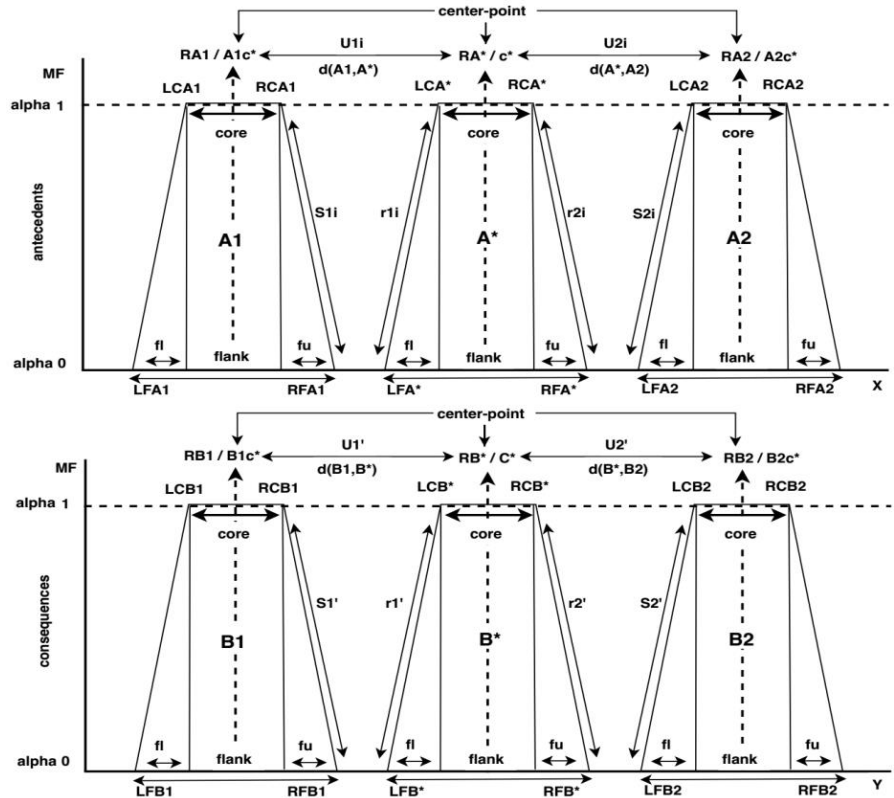

*Figure 3: Fuzzy Interpolation With Trapezoidal Fuzzy Sets (The Antecedent Part And Observation) And (The*

Where $A_1$, $A_2$ refer to the fuzzy sets of the antecedents, $B_1$, $B_2$ denote the consequent fuzzy sets. $A^*$ denotes to the new input (observation), $B^*$ refers to the conclusion. The characteristic points of the trapezoidal membership function could be defined by four values (LF, LC, RC, RF), the (LC, RC) refer to the left and the right core, the (LF, RF) refer the left and the right flank. ($RA_1$, $RA^*$, $RA_2$) denote the center point of the fuzzy sets in antecedents side and similarly the ($RB_1$, $RB^*$, $RB_2$) denote the center points of the fuzzy sets in consequents side, (fl, s2, r1) and (fu s1, r2) denote the left and the right fuzziness for each fuzzy set, ($U_i$, U') denotes the distance between the center points of the fuzzy sets.

## 3. FRI METHODS

There are many fuzzy rule interpolation methods exists, classified into two groups. The first group obtains the conclusion in a single step (directly) and the second group demands two-steps to compute the conclusion, using different algorithms in each step. This section presents an overview about some of the implemented FRI methods.

### 3.1 KH Interpolation Method

The first method which was proposed for FRI is called the KH (linear interpolation) method, this method was published by Kóczy and Hirota [3]. Concerning the common general conditions for FRI methods suggested in [15], The KH rule interpolation method needs the following conditions to be satisfied: the fuzzy sets in both antecedents and consequents must be convex and normal (CNF) with bounded support and at least a partial ordering must exist between fuzzy sets in the universes of discourse.

The conclusion in KH interpolation method produced directly based on the α-cuts of the observation and the fuzzy rule-base, it can be calculated by using the fundamental equation of the KH FRI (1), which is based on the lower and upper fuzzy distances between fuzzy sets [16]. The upper and lower endpoints could be used to calculate the distance between the conclusion and the consequent which must be analogous to the upper and lower fuzzy distances between observation and antecedents.

$$d(A^*, A1):d(A^*, A2) = d(B^*, B1):d(B^*, B2) \quad (1)$$

Where (d) refers to the Euclidean distance that could be used between the fuzzy sets ($A_1$, $A_2$) and ($B_1$, $B_2$).

The conclusion $B^*$ in this method could be calculated based on the lower and upper fuzzy distances between the fuzzy sets of the antecedents, consequent and observation. Figure (3) illustrates the main points (core and flank) of the trapezoidal fuzzy sets which could use in order to compute the conclusion $B^*$ as follows:

The right (core) can be calculated by the Equation (2):

$$RCB^* = \frac{d_1 RC \times RCB_1 + d_2 RC \times RCB_2}{d_1 RC + d_2 RC} \quad (2)$$

Where

$$d_1 RC = \sqrt{\sum_{i=1}^{k}(RCA_i^* - RCA_{i1})^2}$$

$$d_2 RC = \sqrt{\sum_{i=1}^{k}(RCA_{i2} - RCA_i^*)^2}$$

And the right (flank) can be calculated by Equation (3):

$$RFB^* = \frac{d_1 RF \times RFB_1 + d_2 RF \times RFB_2}{d_1 RF + d_2 RF} \quad (3)$$

$$d_1 RF = \sqrt{\sum_{i=1}^{k}(RFA_i^* - RFA_{i1})^2}$$

$$d_2 RF = \sqrt{\sum_{i=1}^{k}(RFA_{i2} - RFA_i^*)^2}$$

The left (core) and the (flank) can be obtained similarly to the above Equation (2 and 3).

The KH method was developed for a single dimension and multi-dimensional antecedent universes as appearing in the previous Equations. The most significant advantage of the KH interpolation is its simplicity and its low computational complexity. However, the disadvantage of this method is the abnormality in the conclusion can be seen in some cases such as in [17], [18], where the lower (left) end of the α-cut interval has a higher value than its upper (right) end point.

### 3.2 The KH Stabilized Method

Many studies introduced a modification of the original KH method to improve the abnormality and to take more than two rules throughout the determination of the conclusion, the extended method was developed to handle and decrease the abnormality of the original KH method is called KH Stabilized that was proposed by Tikk, .et.al. [5].

The main idea of this method is to take all flaking rules of the observation which is getting better with the growth of the number of the rules taken into consideration to conclude the conclusion, using the extent of the inverse distance of the antecedents and observation of fuzzy sets. The universal approximation property holds if the distance function is raised to the power of the inputs dimension.

The authors of [5] propose using formulas to calculate the upper and lower endpoints of α- cuts of the approximated consequence which contain the distance on the $n^{th}$ power as shown via the Equations (4 and 5):

$$\min B_\alpha^* = \frac{\sum_{i=1}^m \frac{\inf(B_{i\alpha})}{d_L^N(A_\alpha^*, A_{i\alpha})}}{\sum_{i=1}^m \frac{1}{d_L^N(A_\alpha^*, A_{i\alpha})}} \quad (4)$$

$$\max B_\alpha^* = \frac{\sum_{i=1}^m \frac{\sup(B_{i\alpha})}{d_U^N(A_\alpha^*, A_{i\alpha})}}{\sum_{i=1}^m \frac{1}{d_U^N(A_\alpha^*, A_{i\alpha})}} \quad (5)$$

The simplest of the KH Stabilized method is the linear interpolation of two rule-bases for the area between their antecedents. In addition, this method can be applied if the observation position is located between two closest rules or hits outside rule-bases.

### 3.3 VKK Interpolation Method

This method was proposed by Vas, Kalmar and Kóczy [4]. The main idea of this method is based on the *center point* and *width ratio*, the conclusion could be calculated by the *center point* and *width ratio* between the antecedent, consequent, and observation fuzzy sets.

The *center point* of the conclusion can be obtained by Equations (6, 7 and 8):

$$Center(B^*) = \frac{LeftCenter \times RightCenter}{d(A_{i1\alpha}, A_{i2\alpha})} \quad (6)$$

$$LeftCenter = d(A_\alpha^*, A_{i2\alpha}) \times Center(B_{i1\alpha}) \quad (7)$$

$$RighCenter = d(A_{i1\alpha}, A_\alpha^*) \times Center(B_{i2\alpha}) \quad (8)$$

Where

$$Center(A_\alpha) = \frac{\inf(A_\alpha) + \sup(A_\alpha)}{2}$$

The *width ratio* of the conclusion can be calculated by Equations (9, 10 and 11):

$$Width(B^*) = \frac{LfetWidth \times RightWidth}{d(A_{i1\alpha}, A_{i2\alpha}) \times WA^*} \quad (9)$$

$$LeftWidth = d(A_\alpha^*, A_{i2\alpha}) \times Width(B_{i1\alpha}/WA_{1i}) \quad (10)$$

$$RightWidth = d(A_{i1\alpha}, A_\alpha^*) \times Width(B_{i2\alpha}/WA_{2i}) \quad (11)$$

Where

$$Width(A_\alpha) = \sup(A_\alpha) - \inf(A_\alpha)$$

The $d(A_{i1\alpha}, A^*_\alpha)$, $d(A^*_\alpha, A_{i2\alpha})$ and $d(A_{i1\alpha}, A_{i2\alpha})$ refer to the distance between antecedents fuzzy sets, the geometric average of the width values metrical is represented by ($WA_{1i}$), ($WA_{2i}$), and ($WA^*$) between the antecedents and observation.

The disadvantage of this method is the abnormality can be appeared in some cases. Nevertheless, the VKK method is distinguished by a low complexity compared to the KH method due to the calculation of the conclusion directly through the center and the width of the fuzzy sets. It is also simple and can be used in several applications without complications.

### 3.4 MACI Interpolation Method

Another method of the FRI called the Modified α-Cut based Interpolation (MACI) method was proposed by Tikk and Baranyi [6]. The main idea of this method is based on the vector's description of the fuzzy sets for eliminating the abnormality problem in the conclusion. The fuzzy set in this method could be described by two vectors space, it can represent the Left and the Right flank of the α-cut levels where the abnormal consequent set is excluded.

The characteristic points that are used in vector description can be represented by the piecewise linear shape of the fuzzy sets where ($a_{-1}$, $a_0$) describe the left flank, and ($a_0$, $a_2$) represent the right flank, also $a_0$ refers to the reference point of the fuzzy set, the Cartesian axes can be represented by $Z_0$, $Z_1$ as shown in Figure (4).

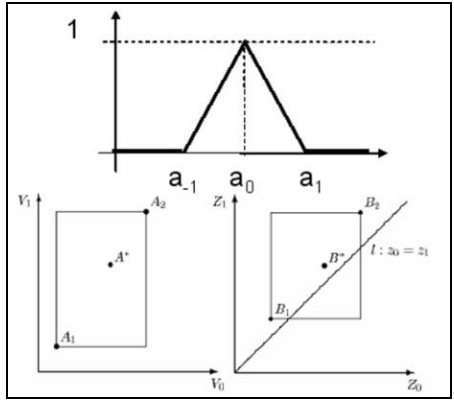

*Figure 4: The Vectors Description Input And Output Fuzzy Sets [6].*

The conclusion in this method could be determined by the transformation of the current characteristic points to a new Cartesian to calculate the conclusion, then, transforming back to the original Cartesian to show the result that could be computed by using the following Equations (12, 13 and 14):

The new Cartesian can be calculated by the vector form:

$$b = [b_0, b_1] \text{ and } b' = [b_0', b_1'] \qquad (12)$$

$$b_0' = b_0 . \sqrt{2} \text{ and } b_1' = b_0 - b_1 \qquad (13)$$

The vector description can be represented by the matrix:

$$b' = bT \qquad (14)$$

Where

$$T = \begin{bmatrix} \sqrt{2} & 0 \\ -1 & 1 \end{bmatrix}$$

The MACI method concentrates on the characteristic points of the fuzzy set ($A_1$, $A^*$ and $A_2$) and the consequents ($B_1$ and $B_2$). It can be described by vectors which involve computing the center point of the conclusion $RB^*$ as shown in Figure (3). The conclusion could be calculated by Equation (15) as follows:

$$RB^* = (1 - \lambda_{core})RB_1 + \lambda_{core} RB_2 \qquad (15)$$

Where

$$\lambda_{core} = \frac{\sqrt{\sum_{i=1}^{k}(RA_i^* - RA_{i1})^2}}{\sqrt{\sum_{i=1}^{k}(RA_{i2} - RA_{i1})^2}}$$

Where, the $RA^*$, $RA_1$, and $RA_2$ denote the reference point of the observation and antecedents fuzzy sets. After computing the conclusion could be transformed back to the original Cartesian by the vector form by applying the Equations (16, 17 and 18):

$$b_0^* = b_0^{*'} . \sqrt{2} \qquad (16)$$

$$b_1^* = b_1^{*'} + b_0^* = b_1^{*'} + (b_0^{*'} / \sqrt{2}) \qquad (17)$$

$$b^* = b^{*'} . T^{-1} \qquad (18)$$

Where

$$T = \begin{bmatrix} 1/\sqrt{2} & 0 \\ 1/\sqrt{2} & 1 \end{bmatrix}$$

For more detailed description of MACI function can be found in [19], [20].

The main advantage of the MACI method is that the conclusion is always a convex and normal fuzzy set. It can also apply multi-dimensional antecedents [6]. On the other hand, the disadvantage of this method (in some instances) is that it does not keep the piecewise linearity of the membership functions.

### 3.5 CRF Interpolation Method

This method was proposed to modify the fuzziness term and to improve α-cut levels. The main idea of this method was introduced in [21] which was called GK method, also the modified version of the GK called the KHG method was published by Kóczy, Hirota, and Gedeon in [7]. The current modified version is called the conservation relative fuzziness (CRF) which follows fundamental equation (FEFRI) (1). This method aims to obtain the conclusion based on determining the core and fuzziness of antecedents, consequents and observation fuzzy sets, the core $c^*$ could be described by ($A_1c^*$, $A_2c^*$, $A^*c^*$) and ($B_1c^*$, $B_2c^*$) as shown in Figure (3), the core of the conclusion could be calculated by using the distances between the antecedents and observation as $d(A^*, A_1)$ and $d(A_2, A^*)$, also between the consequents fuzzy sets $d(B_1, B_2)$.

In addition, the fuzziness of the conclusion could be determined by calculating the (A₁fU, A*fL) that must have the same fuzziness of the (B₁fU, B*fL), and similarly the fuzziness between (A*fU, A₂fL) and (B*fU, B₂fL) as shown in Figure (3).

The core of the conclusion C* can be calculated by Equation (19):

$$C^* = c^* \times \frac{d_1(B_1, B_2)}{d_1(A_1, A_2)} \qquad (19)$$

Where c* denotes the core of the observation, and d1 denotes the distance between $A_1$ and $A_2$ which can be calculated as follows:

$$d_1 = (A_1, A_2) = A_2 c^* - A_1 c^*$$

$$d_1 = (B_1, B_2) = B_2 c^* - B_1 c^*$$

The general fundamental Equation (1) can be applied to determine the distance between the current fuzzy sets through Equation (20):

$$\frac{d_1(A_1, A^*)}{d_1(A^*, A_2)} = \frac{d_1(B_1, B^*)}{d_1(B^*, B_2)} \qquad (20)$$

So, Equation (20) can be used to calculate the core of the conclusion by the distance of the following Equation (21 and 22):

$$d_1(B_1, B^*) = \frac{d_1(A_1, A^*) \times d_1(B_1, B_2)}{d_1(A_1, A_2)} \qquad (21)$$

And similarity,

$$d_1(B^*, B_2) = \frac{d_1(A^*, A_2) \times d_1(B_1, B_2)}{d_1(A_1, A_2)} \qquad (22)$$

Where the distance between the fuzzy sets can be computed as the following formula:

$$d_1(A_1, A_2) = \sqrt{\frac{(A_{2c^*} - A_{1c^*})^2}{range^2}}$$

The fuzziness of the conclusion can be determined by the left and the right flanks by the current fuzzy sets as follows by Equations (23 and 24):

$$B_{fL}^* = A_{fL}^* \times \frac{B_{1fU}}{A_{1fU}} \qquad (23)$$

$$B_{fU}^* = A_{fU}^* \times \frac{B_{2fL}}{A_{2fL}} \qquad (24)$$

Accordingly, the equations in [7], it is possible to compute the A*fL, A*fU, A₁fU, A₂fL, B₁fU, and B₂fL which are based on the calculation of the (inf) and (sup) of the current fuzzy sets.

The previous Equations (19-24) of the CRF method were introduced to be applied by single dimensional input, and also, it can be applied in multi-dimensional input by using the expression in [7].

The advantage of this method is that the flanks are used to define the conclusion, therefore, this method can be applied arbitrarily on fuzzy set shapes. Additionally, the observation position must be surrounding two rule-bases s to get a conclusion.

### 3.6 IMUL Interpolation Method

This method was proposed by Wong, Gedeon and Tikk [8], the IMUL is introduced to avoid the abnormal conclusion and improve the multidimensional α-cut (levels). This method was presented to combine the features of MACI method [6] and Conservation of Relative Fuzziness (CRF) method [7].

The IMUL method applied the vector description, it can describe the characteristics points of the fuzzy sets through advantageous the transformation feature of MACI method, and representing the fuzziness of the input and output by CRF method. The conclusion could be calculated between the characteristic points of the antecedent fuzzy sets which are neighboring to the observation as shown in Figures (3).

The conclusion in IMUL method is based on calculating the reference point (RB*) and the left / right core (LCB*, RCB*), the reference point could be computed by Equation (8). The left and right core can be calculated through the following Equations (25 and 26):

The right core:

$$RCB^* = (1 - \lambda_{right})RCB_1 + \lambda_{right} RCB_2 + (\lambda_{core} - \lambda_{right})(RB_2 + RB_1) \qquad (25)$$

where

$$\lambda_{right} = \frac{\sqrt{\sum_{i=1}^{k}(RCA_i^* - RCA_{i1})^2}}{\sqrt{\sum_{i=1}^{k}(RCA_{i2} - RCA_{i1})^2}}$$

The left core:

$$LCB^* = (1 - \lambda_{left})LCB_1 + \lambda_{left} LCB_2 + (\lambda_{core} - \lambda_{left})(RB_2 + RB_1) \quad (26)$$

where

$$\lambda_{left} = \frac{\sqrt{\sum_{i=1}^{k}(LCA_i^* - LCA_{i1})^2}}{\sqrt{\sum_{i=1}^{k}(LCA_{i2} - LCA_{i1})^2}}$$

The conclusion flanks (LFB*, RFB*) can be computed by following Equation (27):

The left flank:

$$LFB^* = LCB^* - r_k(1 + \left|\frac{S'}{U'} - \frac{S}{U}\right|) \quad (27)$$

Where the LFB* denotes the left flank fuzziness of the conclusion B*, and the LCB* denotes the left core, the right flank can be calculated in the same way, the variables (r, s, u, s', u') are used to determine the fuzziness between the fuzzy sets to calculate the conclusion flank ([8], [19]).

One of the benefits of using IMUL method is that the conclusion can be obtained by computing core and fuzziness focusing on the information of the consequents (outputs) and the information of the antecedents fuzzy sets that are given correct results. Moreover, IMUL method can be applied on single dimension and also in multi-dimensional inputs space (see the examples in [8]).

### 3.7 GM Interpolation Method

The first method in the second group of the interpolation methods which demanded two-steps to get the conclusion is called GM method. It was published by Baranyi et al. [9], the conclusion in this method could be determined by two algorithms. The first one is based on the fuzzy relation and the second one is based on semantics of the relations. The GM method will adopt the characterization of the position fuzzy sets to determine the reference points (core), thus, the distance between the observation and antecedents fuzzy sets can be calculated based on the reference points via Equation (28) instead of using the interpolating α-cut levels.

$$d(A_1, A_2) = |RP(A_2) - RP(A_1)| \quad (28)$$

where $A_1$ and $A_2$ are the fuzzy sets, the reference point is (RP) and (d) denotes the distance of the sets.

The conclusion (interpolation) can be obtained by applying the following primary two steps:
The first step is to generate a new interpolated rule $R^i: A^i \rightarrow B^i$, which is positioned between rules $R_1$ and $R_2$ via Equation (29), the position of the new rule is the same position of the observation, so, each fuzzy set of the antecedents is used to produce the new rule which must be identical with the reference point of the observation fuzzy set in the corresponding dimension.

$$R^i = f^{Interpolation}(R_1, R_2) \quad (29)$$

This step could be divided into three stages:
1. The first stage, a set interpolation technique will help to determine the antecedent set shapes of the interpolated rule.
2. The second stage, the reference points of the observation and the consequent sets can be used to determine the reference point of the conclusion, where the rules could be taken into consideration, for example, using the fundamental equation of the fuzzy rule interpolation (FEFRI) (Equation (1)).
3. The third stage, the shapes of the consequent sets could be determined by the interpolated rule using the same set interpolation technique as (stage 1) as shown in Figure (5).

Many techniques were proposed for this step of the set interpolation technique (e.g. SCM, FPL, etc.). The Solid Cutting Technique (SCM) is introduced for this step, the main idea of this technique is that all the associated sets are rotated by 90° about a vertical axis which is passed through their reference point, then by connecting the similar points of antecedents and consequents, two solids can be constructed: one in the input and one in the output dimension (Figure (5)) where the solid was created in an input dimension could be described.

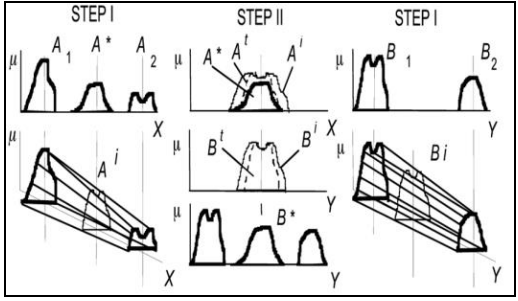

*Figure 5: The Main Steps Of The GM Method [9].*

The second step in the GM method is that the new rule could be specified as a part of the extended rule of the approximate conclusion, as a conclusion of the inference method is defined by determining this rule. In many instances, there is no identical similarity between the rule and the observation part, for this purpose, many techniques are used to handle the mismatch by either the Transformation of the Fuzzy Relation (TFR) technique or by Fixed Point Law (FPL).

This step could be divided to two stages:

The first stage, the Transformation of the Fuzzy Relation (TFR) technique could be applied, where the interrelation function [9] is generated between the observation ($A^*$) and the antecedent ($A^i$) set, there is mapping between observation ($A^*$) and antecedent ($A^i$) by the endpoints of the support and reference point (RP), as shown in Figure (6), the interrelation area can be represented by the endpoints of the supports sets. The purpose of the first phase is to improve the proportion of the area of interrelation mapping between ($A^i$ and $B^i$) sets, it can correspond with the support of the observation and the horizontal side of the square. Hence, the support of the conversion set ($A^t$) is the same support of the ($A^*$), the membership in both cases (At, $A^*$) is the same as its interrelated point in the Antecedents part.

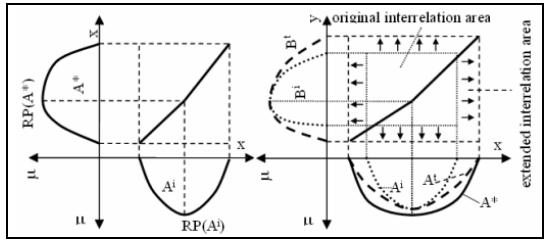

*Figure 6: The Interrelation Functions [9].*

The second stage, the FPL (Fixed Point Law) technique can be used where an interrelation function is created between observation ($A^*$) and the transformed antecedents sets ($A^t$), the (FPL) technique is used to calculate the difference between the membership values for each interrelated point set, this difference can be applied to determine the approximate conclusion from the transformed consequent sets ($B^t$) that will take into consideration the interrelation between transformed ($A^t$) and transformed ($B^t$) [9].

The main advantage of this method (GM) is to avoid the abnormal fuzzy conclusion, there is no restriction to CNF sets and preserving normality, it preserves linearity and is compatible with the rule base, finally it investigates the monotonicity and the continuity.

### 3.8 FRIPOC Interpolation Method

A new fuzzy interpolation method which is based on the reasoning method by using the concept of the linguistic term shifting and polar cut which is called Fuzzy Rule Interpolation based on POlar Cuts (FRIPOC), this method was proposed by Johanyák and Kovács [10], where it is appropriate in case of sparse and dense rule bases. The general formula that can be described to show the reference point which is specified to calculate the interpolated of the Antecedents $RP(A^i_j)$ and the consequent $RP(B^i_l)$ sets which could be calculated by Equation (30).

$$RPB^i_l = f(RP(A^i_1), RP(A^i_2),..., RP(A^i_j),..., RP(A^i_{na}),) \quad (30)$$

This method is based on the position of the fuzzy sets which is characterized by a reference point during the calculations, the reference point $RP(B^i_l)$ can be determined by several techniques Figure (7). The presented technique to determine the reference point can be calculated by Equations (31 and 32). The FRIPOC method essentially follows the GM method [9], where the conclusion can be done by applying two steps: the first step is to define the new rule based on the position of the antecedents part that describes the observation in each dimension, this means the reference point of the observation and antecedents set are identical.

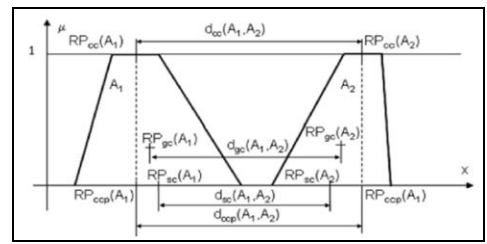

*Figure 7: Choices For The Reference Point And The Associated Set Distances [10].*

$$RP(B_l^i) = \frac{\sum_{j=1}^{N} RP(B_{lj}).s_j}{\sum_{j=1}^{N} s_j} \quad (31)$$

$$s_j = \frac{1}{d(RA^i, RA_j)^2} = \frac{1}{\sum_{k=1}^{na} d(RP(A_k^i), RP(A_{jk}))^2} \quad (32)$$

Where $RP(B^i_l)$ is the RP of the consequent sets, $s_j$ denotes to weight attached to the rule, (l) refers to the number of dimensions, (N) denotes the number of the rules, (j) refers to the actual rule, $RA^i$ and $RA_j$ denote the antecedent rule [10].

Accordingly, the new rule could be determined by two steps: The first step could be described by three stages as follows: 1) the fuzzy sets of the antecedents are estimated by utilizing the set interpolation technique Fuzzy SEt interpolAtion Technique bases on Polar cut (FEAT-p) that is independently in each antecedent dimension, the main purpose of this technique is that the whole sets of the partition are shifted horizontally into the reference point of the observation, i.e. their reference points are identical with the interpolation point. 2) the new fuzzy set is determined based on the polar cut, where the fuzzy set can be specified by using the polar distance of each polar cut level as a weighted mean of the similar polar distances of the forecasted identified sets. 3) the fuzzy set will determine the consequent by FEAT-p technique in the same way as (first stage). Thus, the new fuzzy set can be calculated by following the formula as shown in Equation (33).

$$\rho(A_{j\theta}^i) = \begin{cases} \frac{\sum_{k=1}^{nj} w_{jk}.\rho(A_{jk\theta})}{\sum_{k=1}^{nj} w_{jk}} & , d(A_j^*, A_{jk}) > 0 \\ \rho(A_{jk\theta}) & d(A_j^*, A_{jk}) = 0, k = 1..n_j \end{cases} \quad (33)$$

The second step in FRIPOC method defines the conclusion which is generated by exciting the new rule based on using the Single Rule Reasoning based on polar cuts (SURE-p) technique [10]. The reference point of the interpolated conclusion and the consequent set are identical to the new rule in the current dimension. Figure (8) describes the distance of the polar that can be calculated based on each polar level, the conclusion can be computed by the modified consequents of the interpolated rule using the average differences, where the technique of correction and control could be used to guarantee the efficacy of the new fuzzy set.

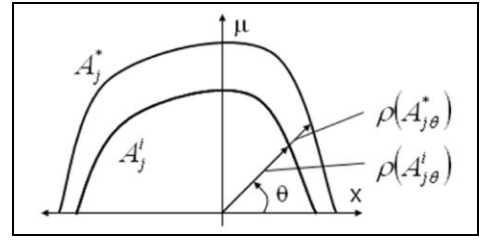

*Figure 8: Polar Distances Utilized For The Estimation Of The Relative Difference [11].*

The main benefits of the FRIPOC method are comprehensibility, the ability to applicability in subnormal cases, and also can be applied if there are no rules surrounding of the observation (extrapolation).

### 3.9 LESFRI Interpolation Method

This method follows the GM method by computing the conclusion based on two steps, this method is called LEast Squares based Fuzzy Rule Interpolation (LESFRI) and was proposed by Johanyák and Kovács Szilveszter [11].

The main idea of this method is the conservation of the weighted average differences measured on the antecedent part, where these modifications could be applied on the consequent side, in which the results usually could be as a set of characteristic points that will not fit with the default shape type of the partition. Therefore, the LESFRI method could be used in order to find the breakpoints of an adequate conclusion. The LESFRI method is based on two-step:

The first step aims to define the interpolation point of the new fuzzy set which can be achieved by three stages as follows [11]:

1. The FEAT-LS technique is used to calculate the antecedent sets for each dimension, where this technique aims to generate a new fuzzy set based on the interpolation points of the fuzzy partitions, thus, all the sets of the partition are shifted horizontally in order to reach the coincidence between their reference points and the interpolation point by using Equation (34).

$$Q_j^L = \sum_{l=1}^{n} w_{lj}.(X_{lj}^L - X_j^L)^2 \quad (34)$$

Where

$$w_j = \frac{1}{d(A_j, A^i)^p}$$

2. The position of the consequent fuzzy sets can be determined for each consequent dimension of the new rule by utilizing a crisp interpolation method by Equation (35).

$$s_l = \frac{1}{d(RA^i, RA_j)^2} = \frac{1}{\sum_{k=1}^{na} d(RP(A_j^i), RP(A_{lj}))^2} \quad (35)$$

3. The characteristic points of the new fuzzy sets shapes are defined by the method of weighted least squares by taking into consideration the similar characteristic points of the overlapped sets which could be used to estimate the conclusion using the observation and the new rule.

The second step in the LESFRI method is the conclusion that could be produced based on the new rule which is required for the calculation of the conclusion because the points of the rule do not fit ideally with the observation in each input dimension. The method that was proposed for this purpose is called SURE-LS as a single rule reasoning method which is based on the α-cut approach. Consequently, all the current antecedent dimensions and consequent fuzzy sets could be described by the break-point α-levels to calculate the conclusion, it must be done independently to the left and right flanks of the fuzzy sets. Additionally, the weighted average of the distances between the endpoints the α-cuts of the rule antecedent and the observation set could be calculated to each side for each level.

The advantages of this method are its capability to produce new linguistic terms that fit into the regularity of the original partitions, as well as its low computational complexity, where it can be applied in case of the interpolation and extrapolation.

**3.10 Scale and Move Interpolation Method**
The scale and move transformation-based method was produced by Huang and Shen [22], it follows the interpolation concept to handle the sparse fuzzy rules. The scale and move method provides the capabilities to work with different fuzzy membership functions types such as (Triangular, Trapezoidal).

The scale and move method is based on the Centre Of Gravity (COG) of the membership functions as shown in Figure (9), this method based on generates a new central rule-base via two neighboring rule-bases that are surrounding the observation.

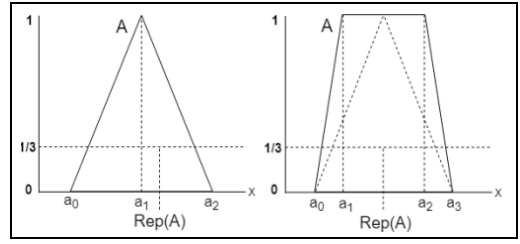

*Figure 9: Representative Value Of A Triangular And Trapezoid Fuzzy Sets [22].*

This scale and move method follows two-steps to obtain the conclusion, the first step is to produce a new central rule-base (A` → B`) is produced within the existing surrounding rule-bases between observation (A*: $A_1 \to B_1$, $A_2 \to B_2$) through to apply the Equation (36):

$$\lambda_{REP} = \frac{d(REP(A_1), REP(A^*))}{d(REP(A_1), REP(A_2))} \quad (36)$$

Where d(Rep($A_1$); Rep($A_2$)) represents the distance between two fuzzy sets $A_1$ and $A_2$. Rep($A_1$) refer to the center of gravity for $A_1$ [22].

The new rule-base (A` → B`) can be calculated by Equations (37 and 38):

$$A' = (1 - \lambda_{REP})A_1 + \lambda_{REP}A_2 \quad (37)$$
$$a_0' = (1 - \lambda_{REP})a_{10} + \lambda_{REP}a_{20}$$
$$a_1' = (1 - \lambda_{REP})a_{11} + \lambda_{REP}a_{21}$$
$$a_2' = (1 - \lambda_{REP})a_{12} + \lambda_{REP}a_{22}$$

$$B' = (1 - \lambda_{REP})B_1 + \lambda_{REP}B_2 \quad (28)$$
$$b_0' = (1 - \lambda_{REP})b_{10} + \lambda_{REP}b_{20}$$
$$b_1' = (1 - \lambda_{REP})b_{11} + \lambda_{REP}b_{21}$$
$$b_2' = (1 - \lambda_{REP})b_{12} + \lambda_{REP}b_{22}$$

The degree of similarity between A` and A∗ is set, it is natural to require that the consequent part B` and B∗ achieve the same similarity degree as follows:

*The more similar X to A`; the more similar Y to B`*

Therefore, the second step is to calculate the A` similarity degree between fuzzy sets (A` and A∗) that is to allow transforming B` to B∗ with the desired degree of similarity by the scale and move. The aim of the Scale transformation is to change the support value of the membership function while keeping its representative value and shape, the aim of the move transformation is to transfer the support of the membership function with keep of its representative.

The advantages of scale and move method that it can handle multiple antecedent variables with simple computation. It guarantees the normality and convexity of the conclusion fuzzy set. It offers the capability to handle the extrapolation issue in direct manner [23]. It preserves the piecewise linearity for interpolations involving arbitrary polygonal fuzzy sets and it uses various definitions for representative values.

## 4. GENERAL DESCRIPTION OF FRI TOOLBOX

The FRI toolbox was developed by Z.C. Johanyák, .et. al. [24] and implemented in MATLAB environment. The main goal of the FRI toolbox is to unify different fuzzy interpolation methods. The general structure of FRI toolbox presented in Figure (10) can run the FRI toolbox and could be used to evaluate the current FRI methods.

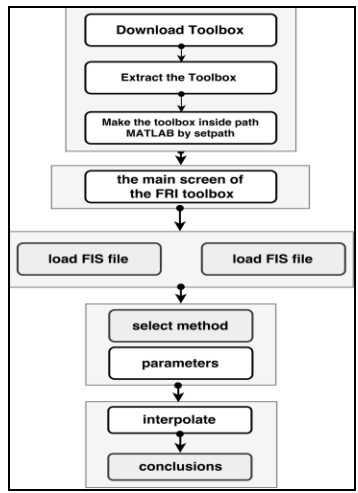

*Figure 10: The General Structure Of FRI Toolbox*

The current version of FRI toolbox is available to download in [26], it includes the following methods (KH, KH Stabilized, MACI, IMUL, CRF, VKK, GM, FRIPOC, LESFRI, and SCALE MOVE). The package of FRI toolbox contains a software with graphical user interface providing an easy-to-use access as shown in Figure (11).

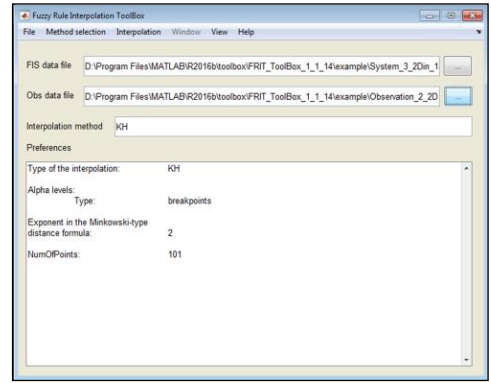

*Figure 11: The Main Panel Of The FRI Toolbox [24]*

In the FRI toolbox, the structure of the fuzzy inference system (FIS) and observation (OBS) were different from the classical inference system. Figure (12) presents an example of FIS within the FRI toolbox. It worths mentioning that, the fuzzy sets have to be convex and normal [3], [25].

```
MF1='A_{1;1}':'trimf',[10 20 30]![0 1 0]
MF2='A_{2;1}':'trapmf',[4.5 5 5.5 6]![0 1 1 0]
MF3='B_{1;1}':'singlmf',[0.46]![1]
```

*Figure 12: The New Parameters Of The Membership Functions That Are Used By The File System In FRI Toolbox*

Where the (trimf), (trapmf) and (singlmf) denote the triangular, trapezoidal and singleton shapes of the fuzzy sets respectively. The $A_{1;1}$, $A_{2;1}$ and $B_{1;1}$ refer to the names of the fuzzy sets of Antecedents and consequent parts. The values [10 20 30], [4.5 5 5.5 6] and [0.46] denote the characteristic points (params) of the fuzzy sets in the universe of the discourse, where the triangular shape takes three values [$a_0$, $a_1$, $a_2$], the trapezoidal shape can be represented by four values [$a_0$, $a_1$, $a_2$, $a_3$], and singleton shape could be described by one value [$a_0$].

The new parameter in FIS general structure is called (paramsy), the characteristic points of the fuzzy sets in case of piecewise linear membership functions as (triangular, trapezoidal, and singleton) that could be represented based on α-cut levels. The lower level will take the value (0) and the upper level will take the value (1). For example, the new parameter of the trapezoidal shape can be represented based on the characteristic points [$a_0$, $a_1$, $a_2$, $a_3$], where the points [$a_0$, $a_3$] refer to the level (0) (lower level) and the points [$a_1$, $a_2$] refer to the level (1) (upper level). Figure (12) describes the new parameter for the trapezoidal membership function (trapmf) which represented by [0 1 1 0].

## 5. NUMERICAL EXAMPLES

In this section, the following (FRI) methods (KH, KH Stabilized, MACI, IMUL, CRF, VKK, GM, FRIPOC, LESFRI, and SCALE MOVE) are presented and discussed in details. The unified numerical examples are applied for the sake of investigating and comparing the FRI methods. These examples selected based on various features, the number of dimensions, the shape of membership functions and the number of membership functions.

Results of these examples would be used to evaluate the FRI methods by following the general conditions of the fuzzy interpolation concept [15], the abnormality and linearity.

These examples were chosen to test the current FRI methods by using FRI toolbox. The triangular, trapezoidal and singleton membership functions are used to describe the antecedent, consequent and observation. Seven examples will be introduced in this section to test the current FRI methods.

The first two examples, single dimension is described the antecedent and consequent, these examples will compare the results based on the difference between the number of the fuzzy sets by using the same membership functions for antecedent, consequent and observation.

The third example will represent the antecedent and consequent by a single dimension, the same number of the fuzzy sets is used for the antecedent, consequent part. This example describes the antecedent, consequent by a different membership function, where the observation represented by the trapezoidal membership function.

The fourth and fifth examples were selected to show the results by using the same membership functions of the antecedents and consequent using different shapes of the observation. These examples are described by using different dimensions, where the antecedent parts are represented by three dimensions and the consequent represented by single dimension.

Table.1 summarizes the unified numerical examples. The antecedents and observations are shown in Figures (13 to 19), the consequents part and conclusions have appeared in Figures (20 to 29).

*Table 1: The Unified Numerical Examples.*

|  | No. Dimensions | | Type of Membership Functions | | | No. of Membership Functions | |
|---|---|---|---|---|---|---|---|
|  | Antecedents | Consequents | Antecedents | Consequents | Observations | Antecedents | Consequents |
| Example 1 | 1 | 1 | **Tri**angular | **Tri**angular | **Tri**angular | 2 | 2 |
| Example 2 | 1 | 1 | **Tri**angular | **Tri**angular | **Tri**angular | 4 | 4 |
| Example 3 | 1 | 1 | **Tri**angular | **Trap**ezoidal | **Trap**ezoidal | 4 | 4 |
| Example 4 | 1 | 1 | **Trap**ezoidal | **Trap**ezoidal | **Sing**leton | 4 | 4 |
| Example 5 | 1 | 1 | **Tri**angular | **Tri**angular | **Sing**leton | 4 | 4 |
| Example 6 | 3 | 1 | **Tri**angular | **Trap**ezoidal | **Tri**angular | 3 | 3 |
| Example 7 | 3 | 1 | **Tri**angular | **Trap**ezoidal | **Sing**leton | 3 | 3 |

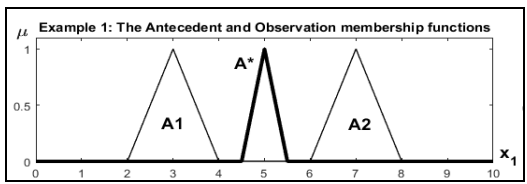

*Figure 13: The Antecedent And Observation Fuzzy Sets For Example (1)*

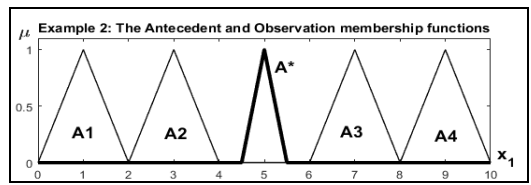

*Figure 14: The Antecedent And Observation Fuzzy Sets For Example (2)*

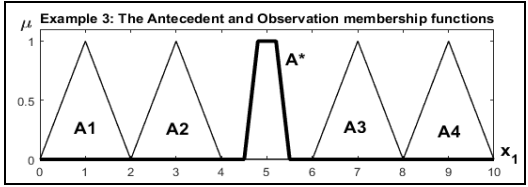
Figure 15: The Antecedent And Observation Fuzzy Sets For Example (3)

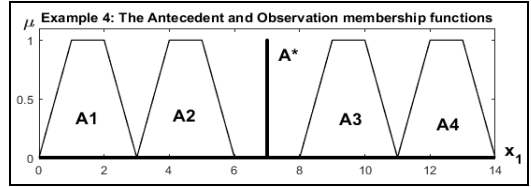
Figure 16: The Antecedent And Observation Fuzzy Sets For Example (4)

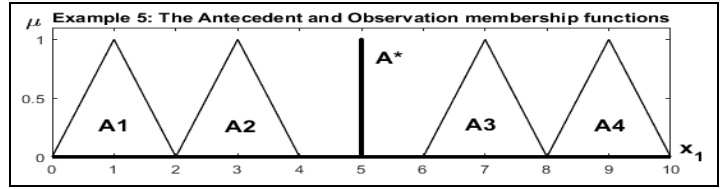
Figure 17: The Antecedent and Observation Fuzzy Sets for Example (5)

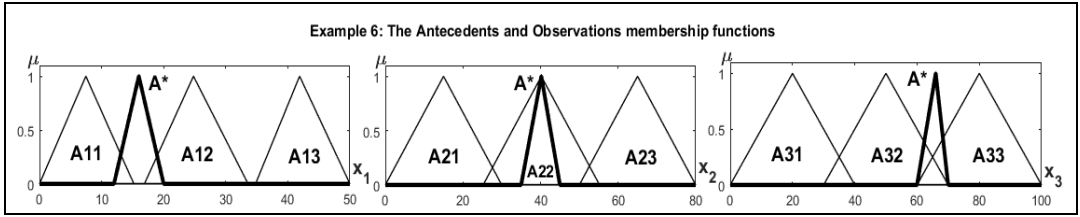
Figure 18: The Antecedents and Observations Fuzzy Sets for Example (6)

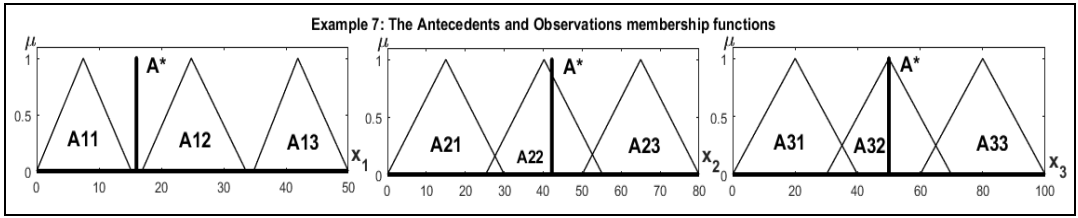
Figure 19: The Antecedents and Observations Fuzzy Sets for Example (7)

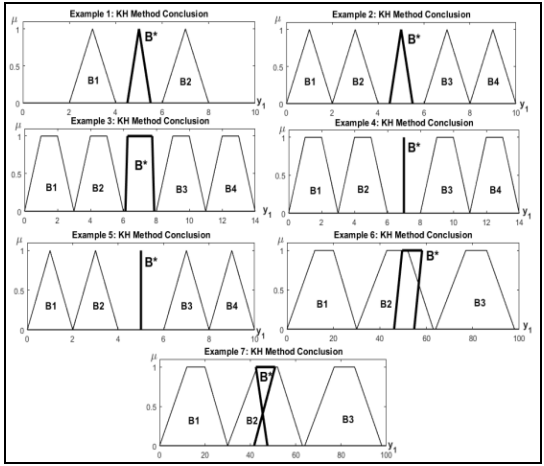
Figure 20: Numerical Examples: KH Conclusions

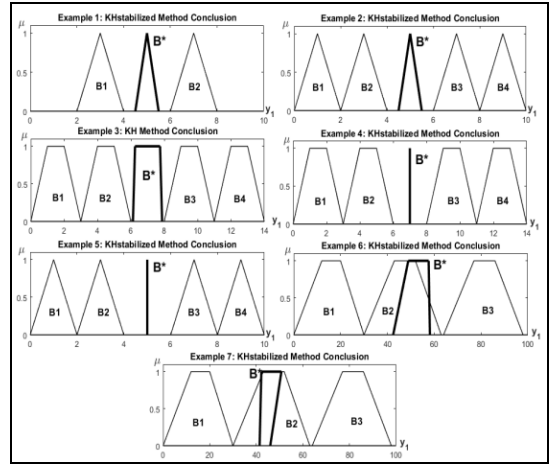
Figure 21: Numerical Examples: KH Stabilized Conclusions

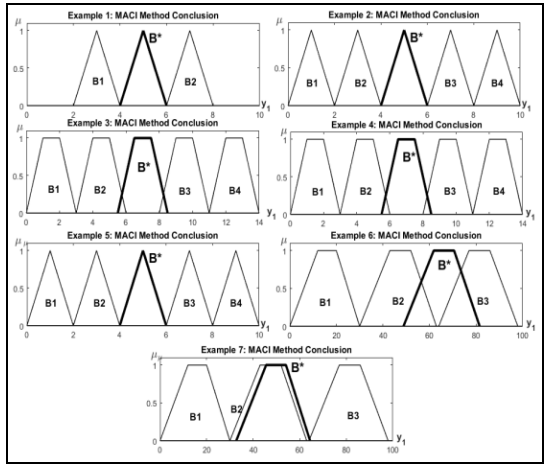

*Figure 22: Numerical Examples: MACI Conclusions*

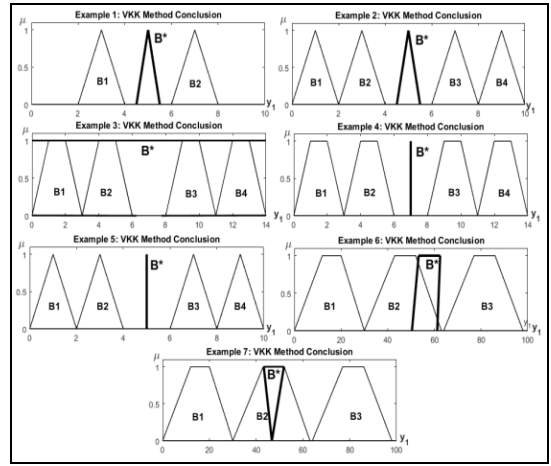

*Figure 25: Numerical Examples: VKK Conclusions*

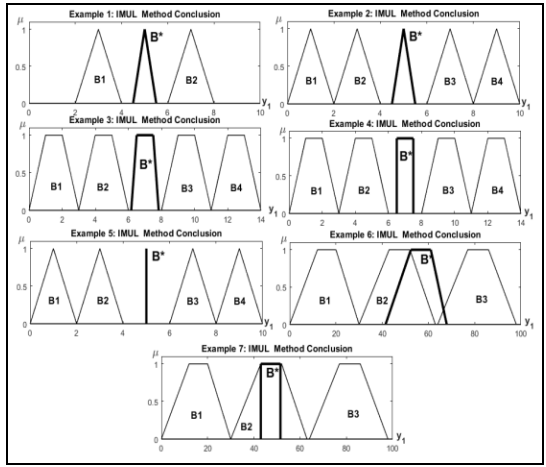

*Figure 23: Numerical Examples: IMUL Conclusions*

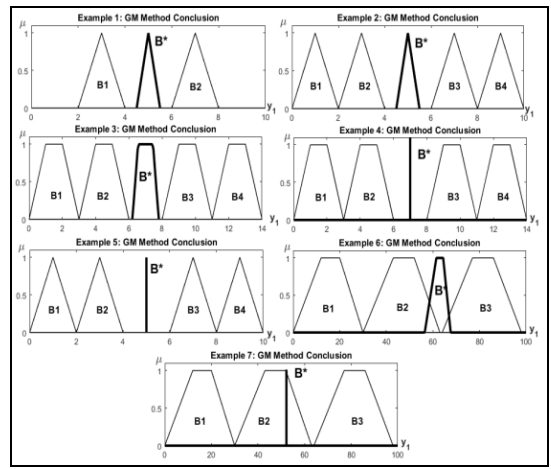

*Figure 26: Numerical Examples: GM Conclusions*

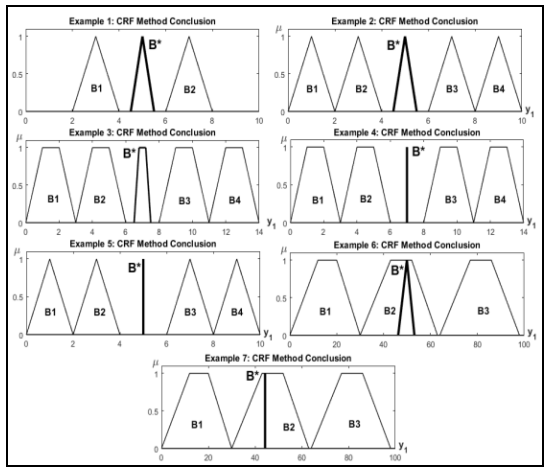

*Figure 24: Numerical Examples: CRF Conclusions*

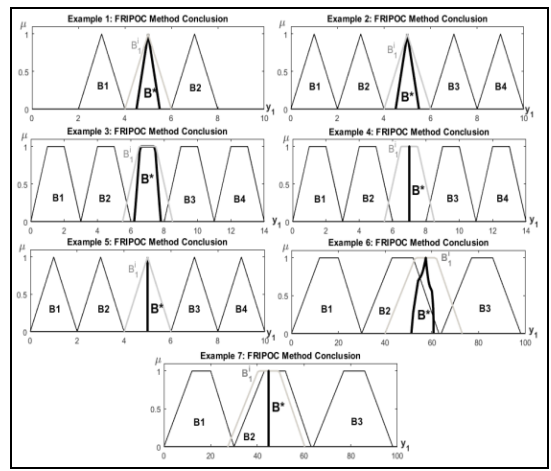

*Figure 27: Numerical Examples: FRIPOC Conclusions*

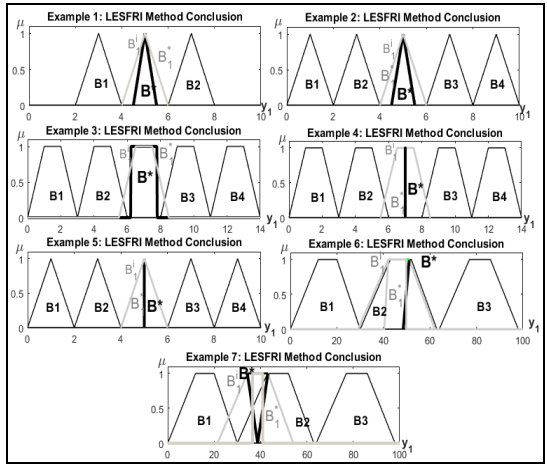

*Figure 28: Numerical Examples: LESFRI Conclusions*

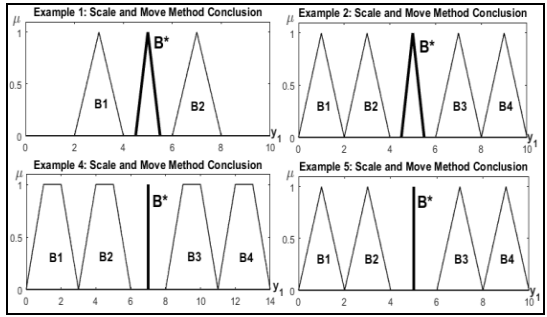

*Figure 29: Numerical Examples: Scale and Move Conclusions*

## 6. RESULTS AND DISCUSSION

The aforementioned results of the numerical examples conclude the following:

According to the antecedents and observations shown in Figures (13,14,16 and 17), the following methods (KH, KH Stabilized, MACI, IMUL, CRF, VKK, GM, FRIPOC, LESFRI, and SCALE MOVE) could be a suitable approach to be implemented as an inference system in a single dimension antecedent, the antecedent and consequent have the same type of membership functions (*Triangular / Trapezoidal*), despite the type of observation membership functions shown in Figures (20 to 29) which illustrated in examples (1,2,4, and 5).

With regard to the antecedents and observations shown in Figure (15), the following methods (KH, KH Stabilized) could be a suitable approach to be implemented as an inference system in a single dimension antecedent, the antecedent and consequent have different type of (*Triangular and Trapezoidal*) respectively, based on the type of the observation shown in Figures (20 and 21) which represented in examples (1,2,4, and 5).

According to the antecedents and observations shown in Figure (15), the following methods (MACI, IMUL, CRF, GM, FRIPOC and LESFRI) could be a suitable approach to be implemented as an inference system in a single dimension, the antecedent and consequent have different type of (*Triangular / Trapezoidal*), regardless of the type of the observation shown in Figures (22, 23, 24, 26, 27 and 28) which illustrated in example (3).

Regarding the antecedents and observations shown in Figures (18 and 19), the following methods (MACI and GM) could be a suitable approach to be implemented as an inference system in multi-dimension antecedents, the antecedent and consequent have different type of membership function (*Triangular / Trapezoidal*), despite the type of observation membership functions shown in Figures (22 and 26) which described in examples (6 and 7).

For antecedents and observations shown in Figures (18 and 19), the following methods (IMUL and CRF) could be a suitable approach to be implemented as an inference system in multi-dimension antecedents, the antecedent and consequent have different type of membership function (*Triangular and Trapezoidal*), regardless of the type of observation membership functions shown in Figures (23 and 24) which defined in examples (6 and 7).

Regarding the antecedents and observations shown in Figure (19), the following method (FRIPOC) could be a suitable approach to be implemented as an inference system in multi-dimension antecedents, the antecedent and consequent have different type of membership function (*Triangular and Trapezoidal*), in case of the type of observation membership function is singleton shown in Figure (27) and defined in example (7).

On the other hand, regarding the antecedents and observations shown in Figure (15), the following method (VKK) suffers from the abnormality in a single dimension antecedent, the antecedent and consequent have different type of (*Triangular and Trapezoidal*) respectively, based on the type of the observation shown in Figure (25) and illustrated in example (3).

Referring to the antecedents and observations shown in Figures (18 and 19), the following methods (KH, KH Stabilized and VKK) suffer from the abnormality in multi-dimension antecedents, whereas the antecedent and consequent have different type of membership function (*Triangular / Trapezoidal*), regardless of the type of observation membership functions shown in Figures (20, 21) and 25) which described in examples (6 and 7).

For to the antecedents and observations shown in Figure (18), the following method (FRIPOC) suffers from the piecewise linearity in multi-dimension antecedents, whereas the antecedent and consequent have different type of membership function (*Triangular / Trapezoidal*), in case of the type of observation membership function is triangular which shown in Figure (27) and displayed in example (6).

Regarding the antecedents and observations shown in Figure (18) and (19), the following method (LESFRI) suffers from the abnormality in multi-dimension antecedents, whereas the antecedent and consequent have different type of membership function (*Triangular and Trapezoidal*), in case of the type of observation membership function is triangular which shown in Figure (28) and defined in example (6 and 7).

## 7. CONCLUSIONS AND FUTURE WORK

This paper contributed to introduce a brief introduction of the extended version of FRI toolbox and how we can use it. In addition, different unified numerical examples introduced to compare between the Fuzzy Rule Interpolation Techniques (FRI) based on the various features especially the shape type of the membership function for the antecedent and consequent, as presented in Table.1.

As a result of the performed examples, KH, KH Stabilized, LESFRI and VKK methods suffer from the abnormality in case of having multi-dimension antecedents and different type of membership functions which described in examples (6 and 7), also, the VKK method suffers from the abnormality in case of having single-dimension which illustrated in example (3). FRIPOC method suffers from piecewise linearity in case of having multi-dimension antecedents and different type of membership functions which displayed in example (6).

In contrast MACI, IMUL, CRF, GM and SCALE MOVE methods did not suffer from abnormality and piecewise linearity according to the unified numerical examples. For future work, we want to take more cases and be standard to find unified examples to compare and evaluate between the FRI methods. Furthermore, the FRI toolbox is still under development by adding new methods.

## ACKNOWLEDGMENTS
The described study was carried out as part of the EFOP-3.6.1-16-00011 Younger and Renewing University - Innovative Knowledge City - institutional development of the University of Miskolc aiming at intelligent specialization project implemented in the framework of the Szechenyi 2020 program. The realization of this project is supported by the European Union, co-financed by the European Social Fund.